\pdfoutput=1

\documentclass[11pt]{article}

\usepackage{acl}
\usepackage{times}
\usepackage{latexsym}
\usepackage[T1]{fontenc}
\usepackage[utf8]{inputenc}
\usepackage{microtype}
\usepackage{inconsolata}
\usepackage{graphicx}
\usepackage{multirow}
\usepackage{soul}
\usepackage{tabto}

\title{cantnlp@DravidianLangTech 2025: A Bag-of-Sounds Approach to Multimodal Hate Speech Detection}

\author{Sidney Wong \\
 Geospatial Research Institute \\
 University of Canterbury \\
 \texttt{sidney.wong@pg.canterbury.ac.nz} \\\And
  Andrew Li \\
  Lake Washington School District \\
  \texttt{landrewi@hotmail.com} \\}

\begin{document}
\maketitle
\begin{abstract}

    This paper presents the systems and results for the Multimodal Social Media Data Analysis in Dravidian Languages (MSMDA-DL) shared task at the Fifth Workshop on Speech, Vision, and Language Technologies for Dravidian Languages (DravidianLangTech-2025). We took a `bag-of-sounds' approach by training our hate speech detection system on the speech (audio) data using transformed Mel spectrogram measures. While our candidate model performed poorly on the test set, our approach offered promising results during training and development for Malayalam and Tamil. With sufficient and well-balanced training data, our results show that it is feasible to use both text and speech (audio) data in the development of multimodal hate speech detection systems.

\end{abstract}

\section{Introduction}
\label{sec:introduction}

    There has been increased recognition within the research field that forms of hate speech on social media are not restricted to written modalities
    of language, but also spoken \cite{chhabra_literature_2023} and non-linguistic (i.e., memes) modalities as well (\citealp{kiela_hateful_2020}). As part of Multimodal Social Media Data Analysis in Dravidian Languages shared task \cite{lal_g_overview_2025}, we propose taking a `bag-of-sounds' approach - analogous to the bag-of-words models - to train our automatic hate speech detection system on the speech (audio) data. We do this by transforming the speech (audio) data into Mel spectrogram measures and training our classification model on the outputs.

\section{Related Works}
\label{sec:related_works}

    The earliest automatic hate speech detection systems relied on different linguistic features such as lexical and syntactic representations \cite{chen_detecting_2012}, template-based and parts-of-speech (POS) tagging \cite{warner_detecting_2012}, topic-modelling \cite{xiang_detecting_2012}, or a combination of lexical, POS, character bigram and term frequency-inverse document frequency (Tf-idf) representations \cite{dinakar_common_2012}. With a focus on hate speech in English, the model performance of these early systems yielded moderate results with limited applications to other language conditions \cite{jahan_systematic_2023}.
    
    The introduction of transformer-based Large Language Models (LLMs), such as BERT \cite{devlin_bert_2019}, saw an increase of word embedding feature representations jointly with neural network models in the development of hate speech detection systems \cite{jahan_systematic_2023}. Hate speech systems are now treated as a text classification task following a standardised pipeline including data set collection and labelling, feature extraction, model learning and development, and evaluation on a multiclass or binary output \cite{rawat_hate_2024}. Both statistical language models and LLMs are used in the development of contemporary state-of-the-art hate speech detection systems.
    
    As with other strands of Computational Linguistics and Natural Language Processing (NLP) for social impact \cite{hovy_social_2016}, there has been a long standing tradition of shared tasks detecting hate speech and offensive language in Indo-Aryan and Dravidian languages (\citealp{chakravarthi_findings_2021}; \citealp{chakravarthi_overview_2022}). The best performing models in \citet{chakravarthi_overview_2024} were developed using open-source multilingual transformer-based LLMs (\citealp{conneau_unsupervised_2020}; \citealp{khanuja_muril_2021}). In addition to testing the usability of transformed-based LLMs in non-English conditions, these systems interrogate the efficacy of text classification in code-switching \cite{yasaswini_iiittdravidianlangtech-eacl2021_2021} and script-switching \cite{wong_cantnlplt-edi-2024_2024} phenomena.
    
    While previous shared tasks have focused solely on written expressions of hate speech, the first multimodal social media data analysis in Dravidian languages (MSMDA-DL) was organised by \citet{chakravarthi_findings_2024} with written and spoken social media language data from YouTube. The shared task provided training data with utterances in two Dravidian languages - Malayalam and Tamil - and annotated for hate speech and abusive language in YouTube videos. Only two systems were submitted as part of \citet{chakravarthi_findings_2024}: \citet{rahman_binary_beastsdravidianlangtech-eacl_2024} and \citet{s_wit_2024}. Of interest to the current paper, \citet{rahman_binary_beastsdravidianlangtech-eacl_2024} extracted Mel-frequency spectrogram and Mel-frequency Cepstral Coefficients (MFCCs) as acoustic features which they incorporated in their ConvLSTM.

    A spectrogram is a visual representation of the acoustic frequency - or the number of vibrations in a sound wave per second - in a speech (audio) signal. The Mel-scale spectrogram, also known as Mel-frequency spectrograms or simply Mel spectrograms, is a transformation of linear machine-readable frequency measures of a spectrogram to a non-linear Mel scale which is the perceptual scale of pitch by human listeners \cite{stevens_scale_1937}. Mel spectrograms and MFCCs \cite{davis_comparison_1980} have been widely used in automatic speaker recognition systems and subjective tasks such as speaker emotion recognition \cite{zhou_exploring_2019}; most recently, these acoustic measures have been included in various forms of NLP classification tasks \cite{arroniz_was_2023}.

    \citet{rahman_binary_beastsdravidianlangtech-eacl_2024} took a Convolutional Neural Network (CNN) with Long-short term memory (LSTM), or ConvLSTM, and a hybrid 3D-CNN with LSTM approach in the development of their multimodal hate speech detection system incorporating visual, audio, and text representations. Although the shared task included training data for three Dravidian languages, only one system was designed for Tamil. During the model development phase, the system achieved a macro average $F_1$-score of 0.71 for Tamil. The system ranked first for Tamil in the shared task with a macro average $F_1$-score of 0.7143 in the test set. \citet{s_wit_2024} did not incorporate the speech (audio) components in the development of their detection system.

\section{Data} 
\label{sec:data}

    \begin{table}
        \caption{\label{tab:train_binary} Binary Class Labels}
        \centering
        \begin{tabular}{cccc}
            \hline
            Class & Malayalam & Tamil & Telugu \\
            \hline
            \textsc{h} & 477 & 227 & 358 \\
            \textsc{n} & 406 & 287 & 198 \\
            \hline
         \end{tabular}
     \end{table}

    \begin{table}
        \caption{\label{tab:train_multiclass} Multiclass Class Labels}
        \centering
        \begin{tabular}{cccc}
            \hline
            Class & Malayalam & Tamil & Telugu \\
            \hline
            \textsc{c} & 186 & 65 & 122 \\
            \textsc{n} & 406 & 287 & 198 \\
            \textsc{p} & 118 & 33 & 58 \\
            \textsc{r} & 91 & 61 & 72 \\
            \textsc{g} & 82 & 68 & 106 \\
            \hline
         \end{tabular}
     \end{table}

     \begin{table*}[]
        \caption{\label{tab:binary_validation} Macro average $F_1$-score on validation training data (Binary).}
        \centering
        \begin{tabular}{ccccccccc}
            \hline
            \multicolumn{1}{l}{} & \multicolumn{2}{c}{\textsc{nb}} & \multicolumn{2}{c}{\textsc{svm}} & \multicolumn{2}{c}{\textsc{lr}} & \multicolumn{2}{c}{\textsc{rf}} \\
             & \textsc{text} & \textsc{speech} & \textsc{text} & \textsc{speech} & \textsc{text} & \textsc{speech} & \textsc{text} & \textsc{speech}  \\
            \hline
            Malayalam & 0.87 & 0.70 & 0.85 & 0.64 & 0.84 & \textbf{0.90} & 0.84 & 0.91 \\
            Tamil & 0.72 & 0.57 & \textbf{0.79} & 0.64 & 0.76 & 0.67 & 0.72 & 0.75 \\
            Telugu & 0.64 & 0.60 & 0.66 & 0.65 & \textbf{0.70} & 0.67 & 0.72 & 0.72 \\
            \hline
         \end{tabular}
     \end{table*}

     \begin{table*}[]
        \caption{\label{tab:multiclass_validation} Macro average $F_1$-score on validation training data (Multiclass).}
        \centering
        \begin{tabular}{ccccccccc}
            \hline
            \multicolumn{1}{l}{} & \multicolumn{2}{c}{\textsc{nb}} & \multicolumn{2}{c}{\textsc{svm}} & \multicolumn{2}{c}{\textsc{lr}} & \multicolumn{2}{c}{\textsc{rf}} \\
             & \textsc{text} & \textsc{speech} & \textsc{text} & \textsc{speech} & \textsc{text} & \textsc{speech} & \textsc{text} & \textsc{speech}  \\
            \hline
            Malayalam & 0.32 & 0.34 & \textbf{0.54} & 0.45 & \textbf{0.54} & \textbf{0.54} & 0.33 & 0.48 \\
            Tamil & 0.14 & 0.21 & \textbf{0.46} & 0.15 & 0.38 & 0.39 & 0.23 & 0.32 \\
            Telugu & 0.24 & 0.28 & \textbf{0.55} & 0.33 & 0.53 & 0.38 & 0.41 & 0.34 \\
            \hline
         \end{tabular}
     \end{table*}

    The training data contained both text and speech (audio) data in three Dravidian languages: Malayalam, Tamil, and Telugu \cite{sreelakshmi_detection_2024}. The target class labels were organised hierarchically including a binary classification with labels `Hate' (\textsc{h}) and `No Hate' (\textsc{n}), and a multiclass classification with five categories included Caste-related hate-speech (\textsc{c}), and Offensive (\textsc{o}), Racist (\textsc{r}), Sexist (\textsc{s}) language, and one residual non-hate speech category (\textsc{n}). The distribution of the target class labels in the training data for the binary classification is presented in Table \ref{tab:train_binary} and for the multiclass classification in Table \ref{tab:train_multiclass}. There is significant class imbalance between target class labels between language conditions and within the training data. In addition to the class labels, the text and speech (audio) observations were identified by subject, binary gender of the speakers, source of utterance, and utterance number. We split the training data set into training and validation set, where the training set with the target labels was used to train the models and the validation set was reserved for performance evaluation.

\section{Methodology}
\label{sec:methodology}

    The primary purpose of the Multimodal Social Media Data Analysis in Dravidian Languages (MSMDA-DL) shared task was to develop a hate-speech detection system that can analyse the text and speech components and predict the respective labels for three Dravidian languages: Malayalam, Tamil, and Telugu. Therefore, we approached this shared task as a classification problem. We trained a suite of candidate multimodal hate speech detection system using a statistical language model approach. While transformer-based LLMs are the state-of-the-art models in hate speech detection \cite{chakravarthi_overview_2024}, there is limited published research testing the use of LLMs in signal processing (i.e., audio data) as existing LLMs, such as BERT \cite{devlin_bert_2019}, are trained on word embeddings from written language data \cite{verma_towards_2024}. This means we cannot directly compare the model performance of text-trained or speech-trained detection systems for the purposes of this shared task. We evaluated the performance of each candidate system according to the macro average $F_1$-score on the training validation data before selecting and submitting the best performing candidate model.
 
\subsection{Data Preprocessing and Feature Engineering} 
\label{subsec:data preprocessing and feature engineering}

    Multimodal hate speech data is a feature of the current shared task. Prior to the model training process, we carried out the following data preprocessing and feature engineering procedures for the two modalities:
 
    \paragraph{Text:} The text data was supplied in the respective Indic (Brahmic) orthographies of each language condition. We applied minimal data preprocessing on the text data as we wanted to preserve the linguistic features between the text and speech (audio) data. \texttt{CountVectorizer} and \texttt{TfidfTransformer} from \texttt{sklearn.feature\_extraction.text} were employed to transform text data into numerical feature vectors suitable for machine learning models. 


    \paragraph{Speech (Audio):} Mel spectrograms were computed for the audio files and converted into decibel units. To ensure uniformity across inputs, all spectrograms were padded to the same shape and reshaped into flat 2D arrays for compatibility with machine learning algorithms. Additional data normalization was implemented to meet the requirements of the Multinomial Na\"ive Bayes algorithm, one of the machine learning algorithms analysed in this research. This process transforms the feature matrix to ensure all feature values are scaled to the range [0,1].

\subsection{Model Training, Evaluation, and Selection Criteria}
\label{subsec:model training, evaluation, and selection criteria}
     
    The training data was split into training and validation sets with a train:test split of 75:25. Four classification models were trained for both binary and multiclass classification tasks across all three Dravidian languages. The binary classification task served as a benchmark. The statistical methods used were as follows: Multinomial Na\"ive Bayes (\textsc{nb}), Linear Support Vector Machine (\textsc{svm}), Logistic Regression Classifier (\textsc{lr}), and Random Forest Classifier (\textsc{rf}). There were 48 candidate models in total according to the following rubric: two classifications \textsc{x} three language conditions \textsc{x} two modalities \textsc{x} four statistical methods.
 
    The performance of each candidate model was evaluated by macro average $F_1$-score. The model performance as measured by macro average $F_1$-score for the binary classification models are presented in Table \ref{tab:binary_validation} and for the multiclass classification models in Table \ref{tab:multiclass_validation}. The best performing model for each language condition (row) is highlighted in \textbf{bold}. For completeness and benchmarking purposes, we have also included the model performance based on $F_1$-score for binary and multiclass classification models in the Appendix as shown in Tables \ref{tab:binary_malayalam} to \ref{tab:multiclass_telugu}.

    For model evaluation, we used a macro average $F_1$ score as the primary metric. Overall, Logistic Regression (\textsc{lr}) achieved the highest macro average $F_1$-score and performed the best among all four algorithms for speech (audio) data in the binary classification candidate models as shown in Table \ref{tab:binary_validation}. In contrast, Linear SVM (\textsc{svm}) had better performance with the multiclass classification candidate models as shown in Table \ref{tab:multiclass_validation}. We notice a significant drop in performance between the binary classification to the multiclass classification models with the maximum macro average $F_1$-score of 0.90 lowering to 0.54.

    Even though the text data trained Linear SVM (\textsc{svm}) models largely outperformed the speech (audio) trained candidate models, we opted for the best performing multiclass speech (audio) data trained models. We justify this decision as the purpose of the shared task was to incorporate multimodal language data and not just one modality. Where text data only encodes linguistic information, we argue speech (audio) data implicitly encodes both linguistic and paralinguistic features of hate speech. Furthermore, the performance of the text trained models Linear SVM (\textsc{svm})  models only performed marginally better than our optimal model - the speech (audio) trained logistic regression (\textsc{lr}) models.

\section{Results}
\label{sec:results}

     \begin{table}[]
        \caption{\label{tab:test_evaluation} macro average $F_1$-score from test evaluation and rank by language.}
        \centering
        \begin{tabular}{lcc}
            \hline
            Language & $F_1$-score & Rank \\
            \hline
            Malayalam & 0.273 & 14 \\
            Tamil & 0.3186 & 9 \\
            Telugu & 0.1774 & 12 \\
            \hline
         \end{tabular}
     \end{table}

     The macro average $F_1$-scores of our candidate model on the test set are presented in Table \ref{tab:test_evaluation}. All three models performed poorly on the test set with a macro average $F_1$-score below chance. In contrast, the best performing model in Malayalam and Tamil was by Team \texttt{SSNTrio} who yielded a macro average $F_1$-score of 0.7511 for Malayalam and 0.7332 for Tamil. The best performing model in Telugu was by Team \texttt{lowes} had a macro average $F_1$-score of 0.3817.
 
\section{Discussion}
\label{sec:discussion}

    \begin{table}
        \caption{\label{tab:test_class} Relative proportion of test ($n=10$) to train data as a percentage (\%) per class label.}
        \centering
        \begin{tabular}{cccc}
            \hline
            Class & Malayalam & Tamil & Telugu \\
            \hline
            \textsc{c} & 5.4 & 15.4 & 8.2 \\
            \textsc{n} & 2.5 & 3.5 & 5.1 \\
            \textsc{p} & 8.5 & 30.3 & 17.2 \\
            \textsc{r} & 11.0 & 16.4 & 13.9 \\
            \textsc{g} & 12.2 & 14.7 & 9.4 \\
            \hline
         \end{tabular}
     \end{table}

    Based on the evaluation metrics alone, our optimal method performed poorly across all three language conditions with a macro average $F_1$-score below chance. With reference to Table \ref{tab:multiclass_validation} and Table \ref{tab:test_evaluation}, we see a significant drop in performance between the validation evaluation and test evaluation. This drop is particularly stark in Malayalam where the macro average $F_1$-score went from 0.54 to 0.273, and for Telugu from 0.38 to 0.1774. Malayalam had a median macro average $F_1$-score of 0.41 and average of 0.38; for Tamil a median of 0.32 and average of 0.34; and for Telugu a median of 0.24 and an average of 0.23.
    
    While the median and mean scores suggest an improvement from \citet{chakravarthi_findings_2024} across the board, we argue the consistently poor performance may indicate there are underlying issues with the training data. One possible explanation for the poor model performance is the class imbalance observed in Table \ref{tab:train_multiclass} where we see not only difference in utterances between language conditions, but also between classes especially in the minority classes. When we consider the relative proportion of utterances in the test evaluation set as shown in Table \ref{tab:test_class}, some utterances in the minority classes are over represented while utterances in the majority classes are under represented.

    As we refer back to the binary classification models as shown in Table \ref{tab:binary_validation} (which were not part of the shared task), we can see that the models performed well not only across all language conditions, but also across the different statistical methods. Even though the models performed poorly on the test set which suggests some modifications are needed to our pipeline, some of the poor model performance can be attributed to the class imbalance in the training and test data. It is possible the decline in performance from the validation to test data suggests possible over-fitting to the training set or other data-set related biases not accounted for in the current model development pipeline which will be worthy of further investigation.

    Possible improvements to our existing model may include further hyper-parameter tuning such as employing optimisation techniques such as GridSearchCV or RandomisedSearchCV to fine-tune parameters for the text classifiers used in our study such as Random Forest, SVM, and Logisitic Regression. We could explore more advanced boosting algorithms like XGBoost, CatBoost, or LightGBM which may improve classification performance. Alternatively, we could look into comparing other speech feature representations such as Mel-Frequency Cepstral Coefficients (MFCCs) in addition to Mel spectrograms which have also been effective in speech-based classification tasks. 

    The current study provides a foundation for future work in the development of multimodal hate speech detection systems. Despite the lower than expected performance of our proposed approach when compared to other teams in the shared task, we demonstrated in this paper that speech data carries valuable extra-linguistic information for hate speech detection. We argue that further improvements in training data representation and model architecture (i.e., with state-of-the-art methodologies) may yield better performance.

\section{Conclusion}
\label{sec:conclusion}

    While our candidate model performed poorly on the test set, our `bag-of-sounds' approach offered promising results during training and development for Malayalam and Tamil. With sufficient and well-balanced training data, our results show that it is feasible to use both text and speech (audio) data in the development of multimodal hate speech detection systems. It is important to note that our current study intentionally avoided state-of-the-art deep learning or large language models to avoid overloading our existing approach with speculative enhancements from deep learning and transformer-based language models; however, we will look to incorporate more sophisticated models, such as ELMo \citep{peters_deep_2018}, in future studies to determine the performance of our proposed approach alongside state-of-the-art methodologies.

\section*{Acknowledgements}

    We would like to thank the four anonymous peer reviewers and the organisers of the Fifth Workshop on Speech and Language Technologies for Dravidian Languages (DravidianLangTech-2025) co-located at the North American Chapter of the Association for Computational Linguistics in Albuquerque, New Mexico. We would also like to acknowledge Fulbright New Zealand | Te Tūāpapa Mātauranga o Aotearoa me Amerika and their partnership with the Ministry of Business, Innovation, and Employment | Hīkina Whakatutuki for their support through the Fulbright New Zealand Science and Innovation Graduate Award.

\bibliography{references.bib}

\appendix

    \begin{table*}[]
        \caption{\label{tab:binary_malayalam} Model comparison metrics by $F_1$-score per class (binary) in Malayalam}
        \centering
        \begin{tabular}{ccccccccc}
            \hline
            \multicolumn{1}{c}{} & \multicolumn{2}{c}{\textsc{nb}} & \multicolumn{2}{c}{\textsc{svm}} & \multicolumn{2}{c}{\textsc{lr}} & \multicolumn{2}{c}{\textsc{rf}} \\
            Class Label & \textsc{text} & \textsc{speech} & \textsc{text} & \textsc{speech} & \textsc{text} & \textsc{speech} & \textsc{text} & \textsc{speech}  \\
            \hline
            \textsc{h} & 0.89 & 0.71 & 0.87 & 0.78 & 0.86 & 0.92 & 0.87 & 0.92 \\
            \textsc{n} & 0.86 & 0.69 & 0.84 & 0.49 & 0.82 & 0.89 & 0.82 & 0.90 \\
            \hline
         \end{tabular}
     \end{table*}

     \begin{table*}[]
        \caption{\label{tab:multiclass_malayalam} Model comparison metrics by $F_1$-score per class (multiclass) in Malayalam}
        \centering
        \begin{tabular}{ccccccccc}
            \hline
            \multicolumn{1}{c}{} & \multicolumn{2}{c}{\textsc{nb}} & \multicolumn{2}{c}{\textsc{svm}} & \multicolumn{2}{c}{\textsc{lr}} & \multicolumn{2}{c}{\textsc{rf}} \\
            Class Label & \textsc{text} & \textsc{speech} & \textsc{text} & \textsc{speech} & \textsc{text} & \textsc{speech} & \textsc{text} & \textsc{speech}  \\
            \hline
            \textsc{c} & 0.72 & 0.44 & 0.80 & 0.73 & 0.68 & 0.70 & 0.70 & 0.63 \\
            \textsc{n} & 0.00 & 0.20 & 0.21 & 0.08 & 0.35 & 0.31 & 0.00 & 0.15 \\
            \textsc{p} & 0.73 & 0.71 & 0.85 & 0.89 & 0.81 & 0.90 & 0.73 & 0.84 \\
            \textsc{r} & 0.14 & 0.19 & 0.38 & 0.55 & 0.42 & 0.58 & 0.13 & 0.45 \\
            \textsc{g} & 0.00 & 0.16 & 0.44 & 0.00 & 0.46 & 0.21 & 0.08 & 0.35 \\
            \hline
         \end{tabular}
     \end{table*}

     \begin{table*}[]
        \caption{\label{tab:binary_tamil} Model comparison metrics by $F_1$-score per class (binary) in Tamil}
        \centering
        \begin{tabular}{ccccccccc}
            \hline
            \multicolumn{1}{c}{} & \multicolumn{2}{c}{\textsc{nb}} & \multicolumn{2}{c}{\textsc{svm}} & \multicolumn{2}{c}{\textsc{lr}} & \multicolumn{2}{c}{\textsc{rf}} \\
            Class Label & \textsc{text} & \textsc{speech} & \textsc{text} & \textsc{speech} & \textsc{text} & \textsc{speech} & \textsc{text} & \textsc{speech}  \\
            \hline
            \textsc{h} & 0.64 & 0.49 & 0.77 & 0.63 & 0.73 & 0.61 & 0.66 & 0.70 \\
            \textsc{n} & 0.80 & 0.64 & 0.81 & 0.65 & 0.78 & 0.73 & 0.77 & 0.79 \\
            \hline
         \end{tabular}
     \end{table*} 

     \begin{table*}[]
        \caption{\label{tab:multiclass_tamil} Model comparison metrics by $F_1$-score per class (multiclass) in Tamil}
        \centering
        \begin{tabular}{ccccccccc}
            \hline
            \multicolumn{1}{c}{} & \multicolumn{2}{c}{\textsc{nb}} & \multicolumn{2}{c}{\textsc{svm}} & \multicolumn{2}{c}{\textsc{lr}} & \multicolumn{2}{c}{\textsc{rf}} \\
            Class Label & \textsc{text} & \textsc{speech} & \textsc{text} & \textsc{speech} & \textsc{text} & \textsc{speech} & \textsc{text} & \textsc{speech}  \\
            \hline
            \textsc{c} & 0.00 & 0.00 & 0.16 & 0.00 & 0.07 & 0.08 & 0.11 & 0.10 \\
            \textsc{n} & 0.00 & 0.15 & 0.47 & 0.00 & 0.49 & 0.35 & 0.32 & 0.21 \\
            \textsc{p} & 0.70 & 0.69 & 0.81 & 0.74 & 0.76 & 0.72 & 0.72 & 0.79 \\
            \textsc{r} & 0.00 & 0.20 & 0.33 & 0.00 & 0.17 & 0.43 & 0.00 & 0.22 \\
            \textsc{g} & 0.00 & 0.00 & 0.53 & 0.00 & 0.39 & 0.36 & 0.00 & 0.26 \\
            \hline
         \end{tabular}
     \end{table*}

    \begin{table*}[]
        \caption{\label{tab:binary_telugu} Model comparison metrics by $F_1$-score per class (binary) in Telugu}
        \centering
        \begin{tabular}{ccccccccc}
            \hline
            \multicolumn{1}{c}{} & \multicolumn{2}{c}{\textsc{nb}} & \multicolumn{2}{c}{\textsc{svm}} & \multicolumn{2}{c}{\textsc{lr}} & \multicolumn{2}{c}{\textsc{rf}} \\
            Class Label & \textsc{text} & \textsc{speech} & \textsc{text} & \textsc{speech} & \textsc{text} & \textsc{speech} & \textsc{text} & \textsc{speech}  \\
            \hline
            \textsc{h} & 0.84 & 0.74 & 0.78 & 0.73 & 0.80 & 0.81 & 0.85 & 0.81 \\
            \textsc{n} & 0.44 & 0.46 & 0.54 & 0.57 & 0.60 & 0.52 & 0.58 & 0.52 \\
            \hline
         \end{tabular}
     \end{table*} 

     \begin{table*}[]
        \caption{\label{tab:multiclass_telugu} Model comparison metrics by $F_1$-score per class (multiclass) in Telugu}
        \centering
        \begin{tabular}{ccccccccc}
            \hline
            \multicolumn{1}{c}{} & \multicolumn{2}{c}{\textsc{nb}} & \multicolumn{2}{c}{\textsc{svm}} & \multicolumn{2}{c}{\textsc{lr}} & \multicolumn{2}{c}{\textsc{rf}} \\
            Class Label & \textsc{text} & \textsc{speech} & \textsc{text} & \textsc{speech} & \textsc{text} & \textsc{speech} & \textsc{text} & \textsc{speech}  \\
            \hline
            \textsc{c} & 0.34 & 0.44 & 0.65 & 0.56 & 0.57 & 0.55 & 0.65 & 0.50 \\
            \textsc{n} & 0.33 & 0.14 & 0.59 & 0.28 & 0.52 & 0.34 & 0.47 & 0.27 \\
            \textsc{p} & 0.51 & 0.43 & 0.65 & 0.53 & 0.67 & 0.49 & 0.68 & 0.57 \\
            \textsc{r} & 0.00 & 0.32 & 0.36 & 0.13 & 0.38 & 0.24 & 0.12 & 0.37 \\
            \textsc{g} & 0.00 & 0.07 & 0.52 & 0.17 & 0.48 & 0.26 & 0.12 & 0.00 \\
            \hline
         \end{tabular}
     \end{table*} 

\section{Limitations}
\label{sec:limitations}

    While we saw promising results in the `bag-of-sounds' approach we proposed in the development of our hate speech detection system; there are two main limitations we wish to address in our system. The first being the use of statistical language models; and secondly, the lack of sociolinguistic input in the development of our model. 
    
    Firstly, our candidate does not use state-of-the-art modelling in the development of our system as we have not incorporated transformer-based LLMs. While we justified our reasons for excluding LLMs in Section \ref{sec:methodology} in order to maintain comparability between the two modalities, we need to determine how we might incorporate the use of LLMs in our system as existing state-of-the-art models rely on multilingual LLMs \cite{chakravarthi_overview_2024}. With the development of LLMs for speech (audio) signal processing \cite{verma_towards_2024}, it may be possible for us to replicate our analysis. As we will discuss in the Ethics Statement, introducing LLMs may inadvertently introduce bias into our hate speech detection system.
    
    This leads us to discuss the second limitation which is the lack of sociolinguistic input in the development of our system. As discussed in Section \ref{sec:data}, each utterance was labelled for one demographic variable - the binary gender classification of the speaker. While we are aware that speech varies based on gender as a result of mechanical and social demographic differences, we have not incorporated in the development of our model. This means it is possible that there are unexplained predictors in the data that are unaccounted for namely acoustic differences between these social categories such as gender, culture, and possibly geographic dialect bias \citep{wong_sociocultural_2024}. Therefore, future work should involve more in-depth analysis on the speech (audio) data which may justify the need to further normalise or standardise the data.
    
\section{Ethics/Broader Impact Statement}
\label{sec:ethic_statement}

    \citet{parra_escartin_ethical_2017} argued that shared tasks play an important role in Computational Linguistics and Natural Language Processing (NLP) as it helps encourage a culture within the field to develop upon the state-of-the-art. With an increased recognition of automatic hate speech detection beyond just written text to other modalities of language \cite{chhabra_literature_2023}, the current shared task plays an important role in how detection can be achieved in not only multimodal but also multilingual contexts.

    In spite of these benefits, there are also ethical issues and negative effects of competition in NLP shared tasks such as secretive behaviour, overlooking the relevance of ethical concerns, unconscious overlooking of ethical concerns, redundancy and replicability in the field among other concerns \cite{parra_escartin_ethical_2017}. With the `datafication' of hate speech an increasing issue within the field of hate speech detection \cite{laaksonen_datafication_2020}, we will consider the ethics and broader impacts of our proposed system within the context of the current shared task guided by the eight principals of \textit{Responsible NLP} \cite{behera_responsible_2023}.

    \paragraph{Principal 1: Well-being} The current system contributes to our current understanding of multimodal automatic hate speech detection in three Dravidian languages. The current system was designed alongside junior researchers which supports development in working with low- and under-resourced language condition often overlooked in NLP research.

    \paragraph{Principal 2: Human-Centred Values} The current system does not include human subjects, external annotators, or additional data from external sources. However, this is also an area of improvement where the researchers can work alongside target communities - namely Malayalam, Tamil, and Telugu speakers - in the development of a system that is fit for purpose.

    \paragraph{Principal 3: Fairness} While we have avoided to the best of our ability to not perpetuate existing prejudice towards marginalised and vulnerable communities, we are aware that hate speech training datasets are sensitive to racial (\citealp{davidson_racial_2019}; \citealp{sap_risk_2019}) and sociocultural (\citealp{lee_hate_2023}; \citealp{wong_sociocultural_2024}). Therefore, we propose that further work is needed to determine the presence of underlying biases within the training data and possible downstream impacts of these biases.

    \paragraph{Principal 4: Privacy and Security} In accordance with the terms and conditions of the shared task, the authors have not re-distributed the data and have only used the data for non-commercial and academic-research purposes. We have not used the data for surveillance, analyses, or research that isolates a group of individuals for unlawful or discriminatory purposes.

    \paragraph{Principal 5: Reliability} We have provided the model performance metrics which can be found throughout the paper and in the Appendix. We acknowledge there will be variances within the metrics due to the stochastic nature of statistical language models. There is limited risk to organisers, authors, or users who wish to reproduce our systems.
    
    \paragraph{Principal 6: Transparency} We have described our system to the best of our ability for other researchers to reproduce our system; however, we are limited to the metadata provided of the training data provided to us by the organisers of the shared task. We have not involved additional human subjects or external annotators.

    \paragraph{Principal 7: Interrogation} We encourage readers to refer to the other system description papers associated with this shared task.

    \paragraph{Principal 8: Accountability}  We encourage readers to contact the authors to discuss the contents of this paper.

\end{document}